%% file: PaperForReview.tex
\documentclass[10pt,twocolumn,letterpaper]{article}

\usepackage[pagenumbers]{wacv} 

\usepackage{graphicx}
\usepackage{amsmath}
\usepackage{amssymb}
\usepackage{bbm}
\usepackage{comment}
\usepackage{booktabs}
\usepackage{multirow,tabularx}

\usepackage[pagebackref,breaklinks,colorlinks]{hyperref}

\usepackage[capitalize]{cleveref}
\crefname{section}{Sec.}{Secs.}
\Crefname{section}{Section}{Sections}
\Crefname{table}{Table}{Tables}
\crefname{table}{Tab.}{Tabs.}

\def\wacvPaperID{2171} 
\def\confName{WACV}
\def\confYear{2024}

\begin{document}

\title{Multi-Source Domain Adaptation for Object Detection \\with Prototype-based Mean-teacher }

\author{Atif Belal$^1$, Akhil Meethal$^1$, Francisco Perdigon Romero$^2$, Marco Pedersoli$^1$, Eric Granger$^1$ \\
$^1$ LIVIA, École de technologie supérieure, Montreal, Canada\\
$^2$ GAIA Montreal, Ericsson Canada\\
{\tt\small \{atif.belal.1, akhil.pilakkatt-meethal.1\}@ens.etsmtl.ca, }
{\tt\small francisco.perdigon.romero@ericsson.com, } \\
{\tt\small \{marco.pedersoli, eric.granger\}@etsmtl.ca }
}
\maketitle

\input{0-abstract}
\input{1-introduction}
\input{2-related-works}
\input{3-proposal}
\input{4-experiments}

\input{5-conclusion}

{\small
\bibliographystyle{ieee_fullname}
\bibliography{egbib}
}
\clearpage

\input{6-supplementary}

\end{document}

%% file: 0-abstract.tex
%
\begin{abstract}
Adapting visual object detectors to operational target domains is a challenging task, commonly achieved using unsupervised domain adaptation (UDA) methods.  Recent studies have shown that when the labeled dataset comes from multiple source domains, treating them as separate domains and performing a multi-source domain adaptation (MSDA) improves the accuracy and robustness over blending these source domains and performing a UDA. For adaptation, existing MSDA methods learn domain-invariant and domain-specific parameters (for each source domain). However, unlike single-source UDA methods, learning domain-specific parameters makes them grow significantly in proportion to the number of source domains. This paper proposes a novel MSDA method called Prototype-based Mean Teacher (PMT), which uses class prototypes instead of domain-specific subnets to encode domain-specific information. These prototypes are learned using a contrastive loss, aligning the same categories across domains and separating different categories far apart. Given the use of prototypes, the number of parameters required for our PMT method does not increase significantly with the number of source domains, thus reducing memory issues and possible overfitting. Empirical studies indicate that PMT outperforms state-of-the-art MSDA methods on several challenging object detection datasets. Our code is available at \href{https://github.com/imatif17/Prototype-Mean-Teacher}{https://github.com/imatif17/Prototype-Mean-Teacher}
\end{abstract}

%% file: 1-introduction.tex
\section{Introduction}
\label{sec:introduction}

{\centering
\begin{figure*}[ht]
\begin{tabular}{ll}
\begin{subfigure}{0.48\textwidth}\includegraphics[width=0.99\columnwidth]{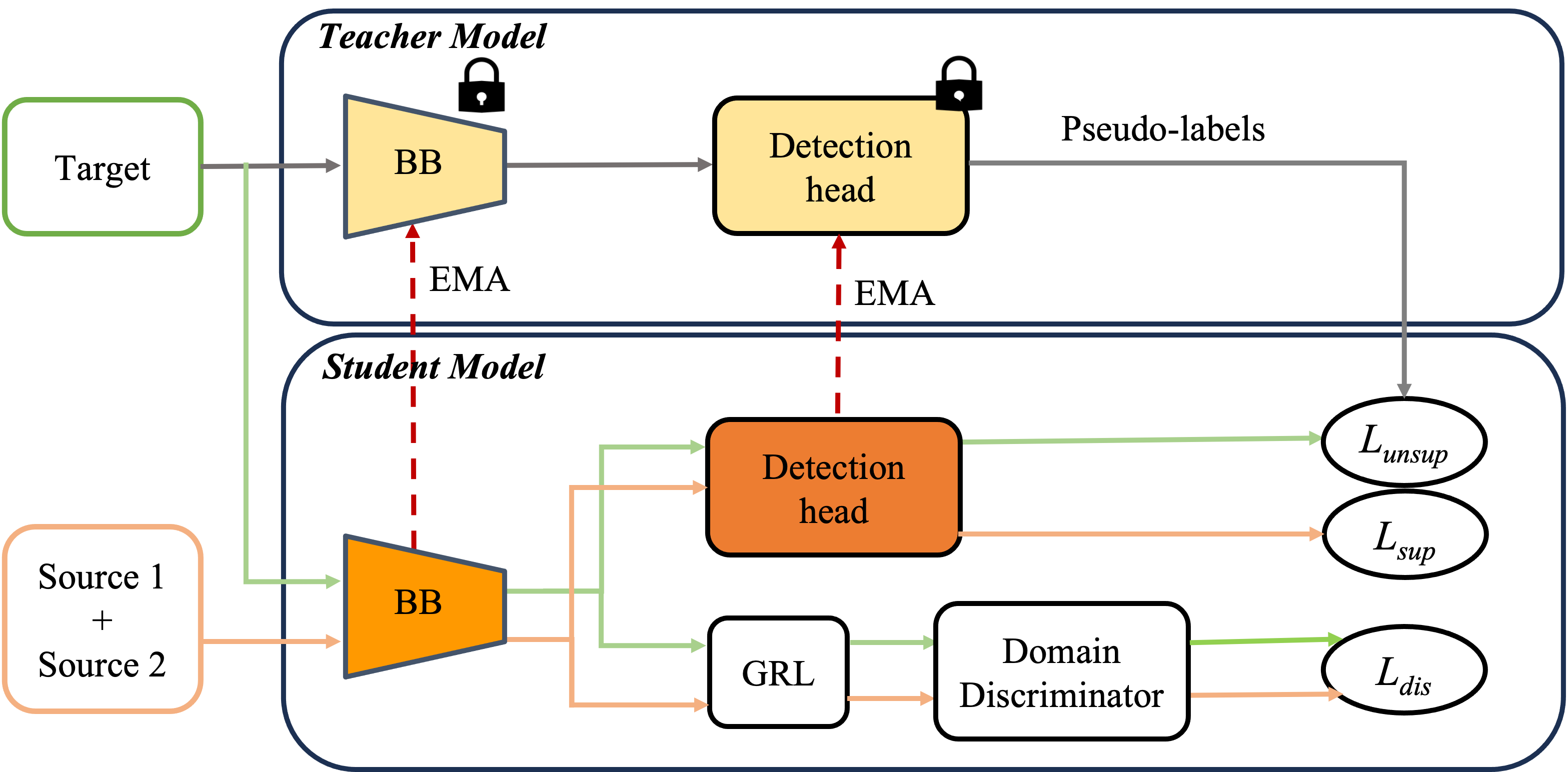}
\vspace{.4in}
\caption{\textbf{Mean-teacher with blending of sources \cite{adaptive_teacher}}. A standard mean-teacher model with a single student and teacher network, trained by blending data from multiple sources. Domain-invariant features are obtained from the backbone trained with a discriminator. No domain-specific information is considered.}
\label{fig:adap_fig}\end{subfigure}&
\begin{subfigure}{0.48\textwidth}\centering\includegraphics[width=0.99\columnwidth]{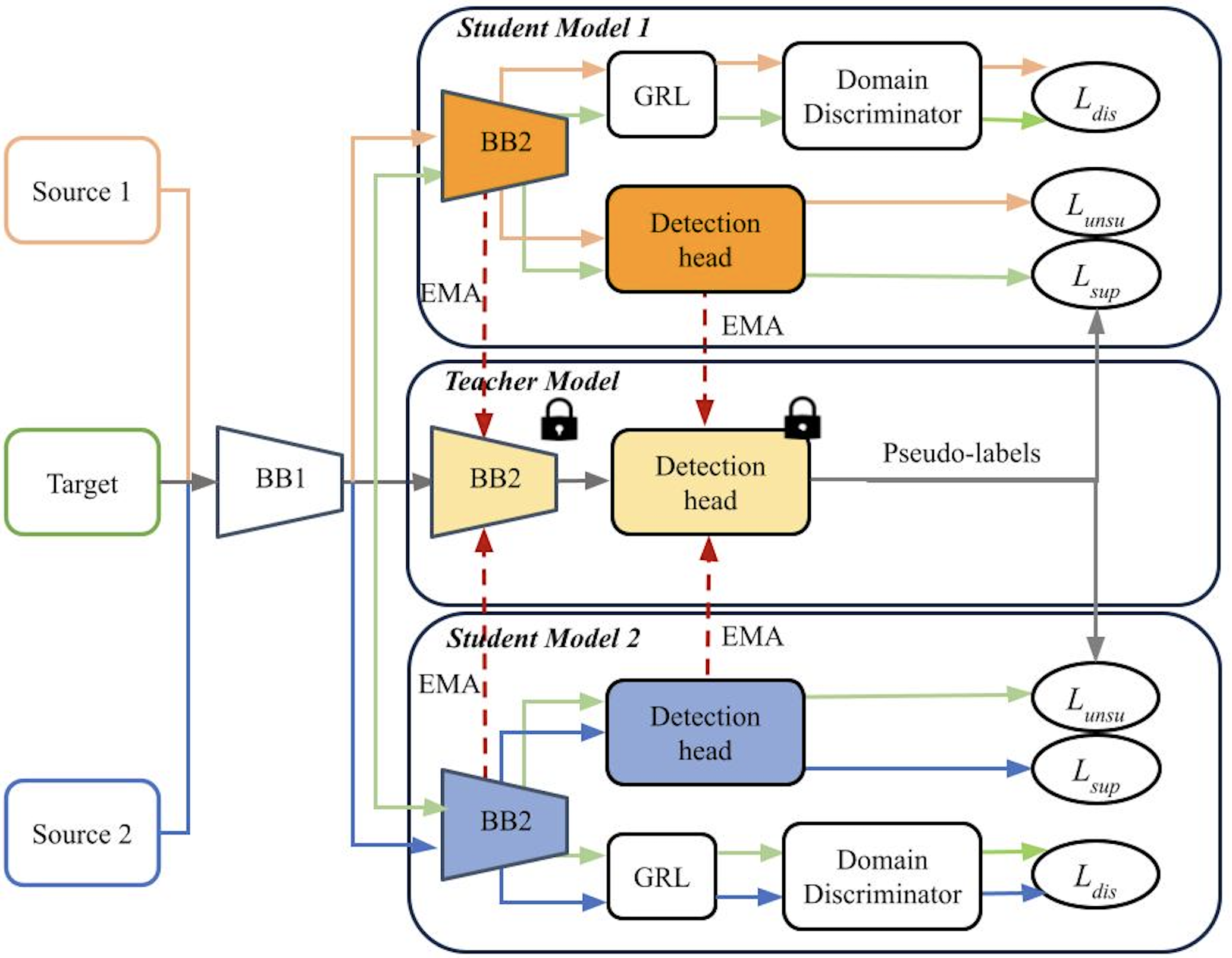}
\caption{\textbf{Divide-and-Merge Spindle Network (DMSN) \cite{dmsn}}. MSDA with multiple student networks and a single teacher network. Domain-invariant features are obtained from the first half of the backbone (BB1). Domain-specific features are obtained from BB2 and the detection head for each source domain.}
\label{fig:dmsn_fig}\end{subfigure}\\
\newline
\begin{subfigure}
{0.48\textwidth}\centering
\includegraphics[width=0.99\columnwidth]{
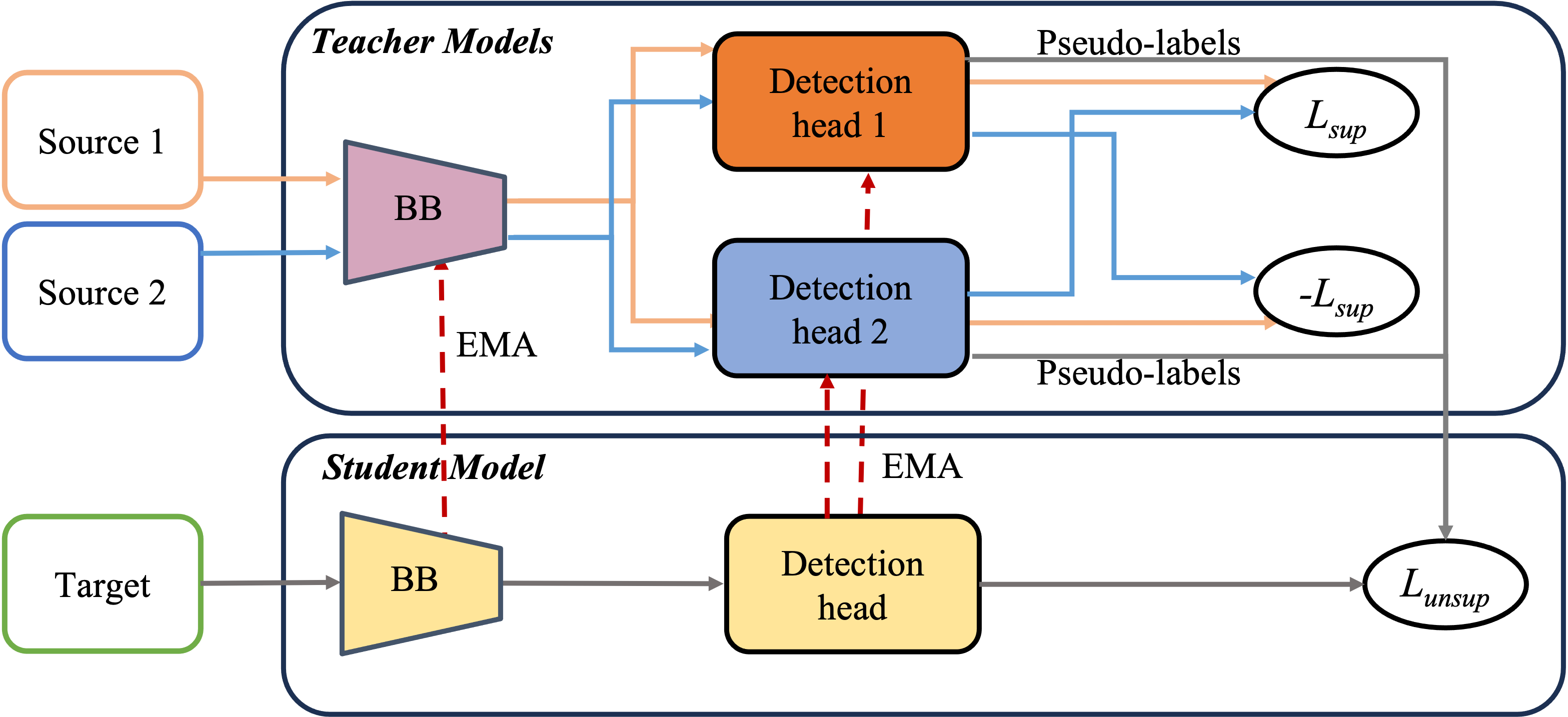}
\vspace{.2in}
\caption{\textbf{Target-Relevant Knowledge Preservation (TRKP) \cite{trkp}}. MSDA with multiple teacher networks and a single student network. Domain-invariant features are obtained from the backbone by reversing detection head gradients. Domain-specific features are obtained from the detection heads.}
\label{fig:trkp_fig}\end{subfigure}&
\begin{subfigure}{0.48\textwidth}\centering
\vspace{.1in}
\includegraphics[width=0.99\columnwidth]{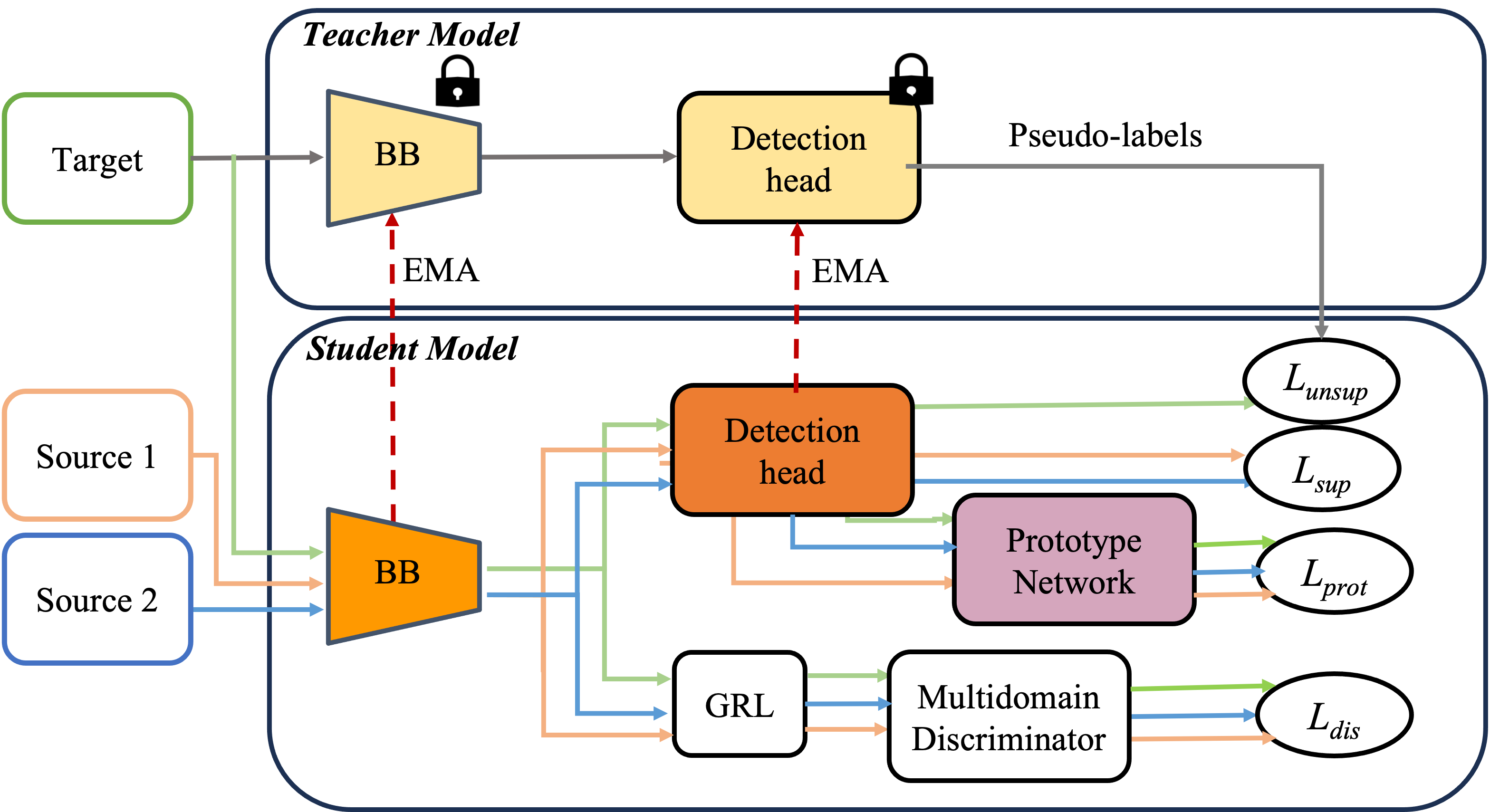}
\vspace{.1in}
\caption{\textbf{Our proposed PMT method}. The original mean teacher framework is used with one student and one teacher. Domain-invariant features are obtained from the backbone trained with a discriminator. Domain-specific information is obtained from domain-specific prototypes for each class.}
\label{fig:our_fig}\end{subfigure}\\
\end{tabular}
\caption{A comparison of the MSDA architectures using the mean-teacher method in the case with two source domains. While state-of-the-art methods require domain-specific parameters (typically a detection head for each source domain) for preserving domain information, our method stores domain-specific information using prototypes for each class and domain.}
\label{tab:mytable}
\end{figure*}
}

Object detection (OD), one of the fundamental tasks in computer vision, has made significant progress in recent years \cite{detic-zhou-2022, detr-Carion-2020, fcos}. However, this progress, typically measured by performance on curated benchmark datasets such as MS-COCO \cite{mscoco-Lin-2014} or Pascal VOC \cite{voc-Everingham-2010}, may decline significantly with changes in the data distribution between training (source) and testing (target) domain data \cite{da_coarse_to_fine-Zheng-2020, da_graph-Xu-2020}. This distribution shift can be caused by several factors, like variations in capture conditions, e.g.,  weather, illumination, resolution, geographic locations, and size of objects.  Since it is too costly to annotate data collected from every operational target domain, researchers have explored a multitude of techniques for unsupervised domain adaptation (UDA), which seeks to find a common representation space between data distributions of a labeled source domain and an unlabeled target domain \cite{da_wild, strongweak, adaptive_teacher}. 

Among the state-of-art UDA strategies for OD, feature alignment \cite{da_wild, selectivesearch} and pseudo-labeling of target data \cite{pseudo_classifier,pseudo_ssd} are the most common. Recently, impressive results were obtained by combining both strategies,  
where the popular mean-teacher \cite{mean_teacher-Antti-2018} framework is used to pseudo-label target domain images, and a gradient reversal layer (GRL) \cite{grl} integrated domain discriminator is used to align features in an adversarial way \cite{adaptive_teacher, MT_graph, MTunbiased}. Mainstream UDA methods rely on a single source domain with labeled data during adaptation. However, in a practical setting, several source datasets may be available, and the datasets may also be collected from multiple domains, e.g., sensors, environment, etc.  The simplest way to use UDA with multiple sources of training data is by blending source domain data to form a single labeled source dataset, 
as shown in \cref{fig:adap_fig}. However, considering each domain as a separate source allows the OD model to address domain discrepancies between individual source distributions explicitly, and aligning sources during adaptation has been found to provide a higher level of recognition accuracy and robustness \cite{MDAN}. This setting is referred to as multi-source domain adaptation (MSDA). 

Two methods have been proposed for MSDA in OD -- Divide-and-Merge Spindle Network (DMSN) \cite{dmsn} and Target-Relevant Knowledge Preservation (TRKP) \cite{trkp}. The overall architecture of both methods can be divided into domain-general and domain-specific model parameters. The initial part of the network learns a domain-invariant representation, while the latter part learns a domain-specific representation. This allows the model to align the features from multiple domains while preserving domain-specific information. As illustrated in \cref{fig:dmsn_fig}, DMSN relies on a mean-teacher training framework, with domain-specific student subnets for each source domain.  
In contrast, as shown in \cref{fig:trkp_fig}, TRKP proposed an adversarial disentanglement method and preserved domain-specific knowledge using a separate detection head for each source domain. The detection results from teacher detection heads are combined to generate pseudo-labels for the student network trained on the target domain.  To fully exploit the potential of multiple sources, both methods advocate for learning domain-specific weights associated with each source domain.  

Learning domain-specific weights for each source domain leads to MSDA architectures that are, however, difficult to train for OD since the number of parameters increases rapidly with the number of sources. It has also been observed that the weighted combination of the source domains based on the heuristics of domain similarities is not optimal, where the similarity of each pair of source-target domains is quantified by a weight value  \cite{msda_prototype_cls-Xu-2022}. Additionally, DMSN and TRKP focus on finding a common representation space between domains through adversarial training \cite{dmsn} and adversarial disentanglement \cite{trkp}, yet ignore the class-wise alignment of object categories. The appearance, scale, orientation, and other visual characteristics of objects may vary across domains. If objects in the source and target domains exhibit significant differences, it may be challenging to align the features extracted by the OD model to accurately detect objects in the target domain. An increase in the number of source domains in MSDA exacerbates this problem. In this paper, we propose learning domain-specific class prototype vectors to address these issues. We argue that MSDA can be simplified without learning domain-specific parameters for each source domain. Instead, learning domain-specific class prototype vectors is sufficient to encode domain-specific information for each source domain. Additionally, given the use of prototypes, the number of parameters of our method does not increase significantly with the number of source domains, thereby reducing memory issues and possible overfitting on the source data.


In this paper, a new MSDA method is introduced for OD that is based on mean-teacher \cite{unbiased_teac} and adversarial training (see \cref{fig:our_fig}). Our Prototype Mean Teacher (PMT) method relies on a cost-effective architecture that stores class and domain-specific information using prototypes. To learn domain-invariant features, a multi-class domain discriminator is integrated into the features generated from the OD backbone. To preserve the domain-specific information of all domains, PMT learns class prototypes using the prototype network for all the domains. Additionally,  the prototypes are trained using a contrastive loss to perform a class-conditional and domain-specific adaptation of the detector. The resulting PMT  method allows adapting detectors to achieve a higher OD accuracy while being conceptually more straightforward than state-of-the-art methods.

\noindent \textbf{Our main contributions are summarized as follows.}\\
(1) A novel Prototype-based Mean Teacher (PMT) method is introduced for the MSDA of ODs. Our approach shows that prototype vectors can elegantly and efficiently encode domain-specific information.\\
(2) Using a multi-domain discriminator and a contrastive loss on the prototypes for each domain and class, the proposed PMT can perform class- and domain-conditional adaptation, improving a detector's performance.\\
(3) Our experimental results and ablations show that the PMT method scales well to the number of source domains and can outperform state-of-the-art MSDA methods for OD on benchmark domain adaptation datasets.

%% file: 2-related-works.tex
\section{Related Works}  
\label{sec:related-work}

Many DL models have been proposed for OD, namely, Faster-RCNN (FRCNN)\cite{fasterrcnn}, Fully Convolutional One Stage (FCOS)\cite{fcos}, You Only Look Once (YOLO)\cite{yolo}, Single Shot Detector (SSD)\cite{ssd}. 
Although our PMT method is independent of the OD model, this paper focuses on FRCNN as our base detector for SOTA comparison. 

\noindent \textbf{(a) Unsupervised Domain Adaptation.} 
UDA methods seek to alleviate the problem of domain shift by adapting the detector trained on a labeled source dataset to a target domain using an unlabeled target dataset. Minimizing the domain discrepancy and adversarial learning are the most popular approaches for UDA. In \cite{da_wild}, the authors integrated a domain discriminator into Faster-RCNN to learn domain-invariant feature representations. Later, \cite{strongweak} proposed strong and weak alignment loss. \cite{progressive} proposed a two-step domain alignment using cycle GAN to mitigate the impact of domain shift between the source and target domains. \cite{crossdomain_proto} proposed a graph-based prototype alignment, using contrastive loss. \cite{adaptive_teacher} utilized a combination of adversarial training and the mean teacher paradigm.  All these methods were designed for UDA from a single source domain. When the source data itself is coming from multiple domains, modeling the domain discrepancy improves the accuracy and robustness of the target OD model \cite{trkp, dmsn}. In our work, we focus on MSDA methods for OD that explicitly consider domain discrepancies among source datasets. 

\noindent \textbf{(b) Multi-Source Domain Adaptation.}
In the MSDA setting, the labeled training data belongs to multiple source domains, and the MSDA method aims to distill the knowledge from these source domains to the target domain.  
Only two methods have been proposed for this setting called DMSN~\cite{dmsn} and TRKP \cite{trkp}. Fundamentally, both methods preserve domain-specific information by utilizing specific subnets for each source domain. Thus the parameters of these methods increase linearly as the number of source domains increases. In our method, we preserve domain-specific information using prototypes for each domain which results in a constant size for the MSDA model regardless of the number of source domains.

\noindent \textbf{(c) Prototype-based Learning.} This learning paradigm is used in different forms for open-world OD \cite{openWOD-Joseph-2021}, semi-supervised OD \cite{ssod_Prot-Li-2022}, domain-adaptive OD \cite{da_Prot-Zhang-2021}, and few-shot OD \cite{fewshot_Prot-Wu-2021}. In open-world OD, prototypes are used to achieve class separation in the feature space and help unknown class identification \cite{openWOD-Joseph-2021}. In semi-supervised OD, prototypes are used for class distribution alignment between pseudo-labels and their highly overlapping proposals \cite{ssod_Prot-Li-2022}. In the detection of objects across domains, prototypes are used to align the foreground and background regions between the source and target domains \cite{da_Prot-Zhang-2021}. Few-shot OD used universal prototypes to learn the invariant object characteristics across seen and novel classes \cite{fewshot_Prot-Wu-2021}. In contrast, we use prototypes in MSDA settings to simplify domain-specific feature learning. Existing MSDA methods learn domain-specific features by using separate networks for each source domain. We show that by using domain-specific prototype vectors instead, we can avoid learning separate networks for each domain.

%% file: 3-proposal.tex
\section{Proposed Method}
\label{sec:proposal}

In the MSDA setting, we assume that there are $N$ source domains $S_1, S_2, …, S_N$ and one target domain $T$. Each source domain can be represented as $S_j = \{(x^j_i, y^j_i)\}^{M_j}_{i=1}$, where $j = 1, 2, ..., N$ indicates the source domain, and where $M_j$ indicates the number of images in the source domain $j$. Here, $x^j_i$ represents the input image $i$ of domain $j$, and  $y^j_i$ represents its corresponding annotations (bounding boxes and object categories). The target domain $T$ can be represented as $T = \{{\{x_i\}^{M_T}_{i=1}}\}$, where $M_T$  indicates the number of images in the target domain. Here, $x_i$ represents the input image $i$ in the target domain. In the work, we assume that all domains have the same $K$ object categories. 

\begin{figure*}[h!]
  \hspace{0.5cm}
  \includegraphics[height=8cm, width=0.94\textwidth]{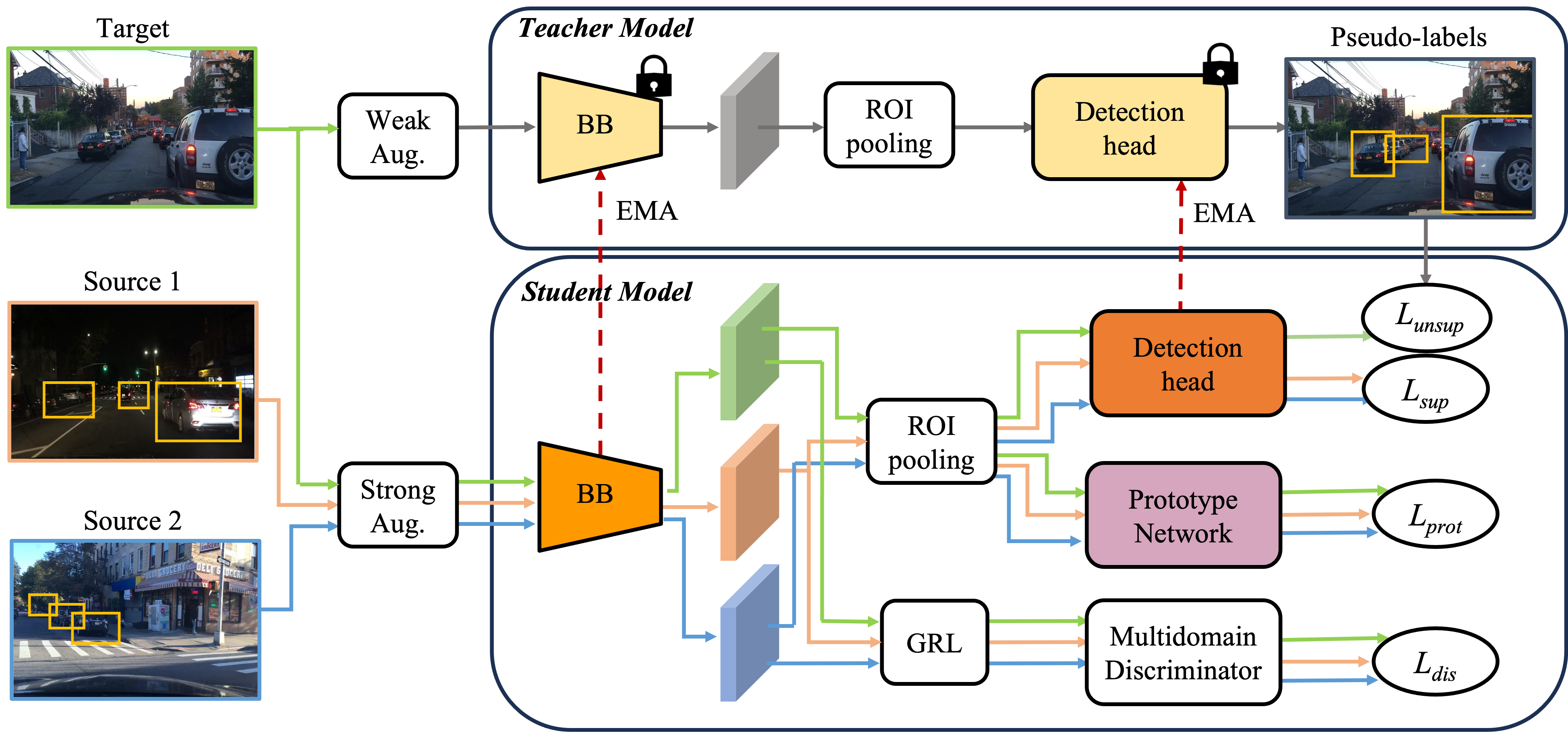}
  \caption{Architectural diagram of the proposed Prototype-based Mean Teacher (PMT) for MSDA. Following the mean-teacher framework, the student is trained with backpropagation, while the teacher is an exponential moving average of the student. The student is trained with images from all domains, and feature alignment is performed at both the image and instance levels using a discriminator and a prototype, respectively. During inference, the teacher model is only employed.}
  \label{fig:multi_source_da_arch}
\end{figure*}

The detailed architectural diagram of our PMT method is illustrated in \cref{fig:multi_source_da_arch}. We use the mean teacher framework with unbiased teacher \cite{unbiased_teac} for semi-supervised OD as our base model. Training of our model is performed in two stages. First, a supervised-only training called the burn-in stage is performed to help achieve reliable pseudo-labels on the target domain because the teacher network is initialized using these weights. Second, the model is adapted with labeled data from the source domains and unlabeled data from the target domain. With the source data, we continue using the supervised training (as in the burn-in stage). Target domain pseudo-labels are obtained from the teacher to train the model on the target domain. To produce domain invariant features, a discriminator is introduced in the network. Additionally, the prototype network is integrated to preserve domain-specific information, and class-conditioned adaptation is performed using a contrastive loss. The rest of this section details the individual components of our proposed PMT. 

\subsection{Supervised Learning}

In our problem setting, the annotations are available for the source domains, so we use them to train the model in a supervised way. For this, we compute the standard supervised detection loss of the object detector: 
\begin{equation}
\label{eq:super}
     \mathcal{L}_{\mbox{sup}} = \sum_{j=1}^{N}\sum_{i=1}^{M_j}\mathcal{L}_{\mbox{cls}}(x^j_i, y^j_i) + \mathcal{L}_{\mbox{reg}}(x^j_i, y^j_i)
\end{equation}
where $\mathcal{L}_{\mbox{cls}}$ is the cross-entropy loss used for bounding box classification, and $\mathcal{L}_{\mbox{reg}}$ is the smooth-L1 loss used for the bounding box regression as in FRCNN. 

\subsection{Learning with Pseudo-labels}

Since annotations are unavailable for the target domain data, we cannot directly proceed with supervised training as with the source domain. We obtain pseudo-labels for images for the target domain using the mean-teacher method \cite{mean_teacher-Antti-2018}. For that, two augmented versions of the target-domain images are created, called weak and strong augmentations. The weak augmentation is simply the image rescaling and horizontal flip transformation. The strong augmentation includes color jittering, grayscale, Gaussian blur, and cutout patches, which perform only pixel-level transformations. We followed the scale ranges provided in \cite{unbiased_teac} for the strong augmentation. Then, the weak and strong augmented versions are processed by the teacher and student networks, respectively, as shown in \cref{fig:multi_source_da_arch}. The predictions made by the teacher model are used as pseudo-labels for the training of the student model. To avoid noisy pseudo-labels from the teacher model, confidence thresholding is applied to the teacher network prediction while computing the pseudo-labels. For this, we followed the same settings as in \cite{unbiased_teac}. The used unsupervised training loss for the target domain using labels from the teacher model is:
\begin{equation}\label{eq:unsuper}
	\begin{aligned}
	\mathcal{L}_{\mbox{unsup}} =\sum_{i=1}^{M_T}\mathcal{L}_{\mbox{cls}}(x_i, \widetilde{y}_i)
	\end{aligned}
\end{equation}
where $\widetilde{y}_i$ represents the filtered pseudo-labels generated by the teacher model. Note that the regression loss is not used here, as the confidence
thresholding is not sufficient to filter the pseudo-labels that are potentially incorrect for bounding box regression \cite{unbiased_teac}.

\subsection{Domain-invariant Features with Discriminator}
The feature representation of a model consists of two parts: domain-invariant and domain-specific. In a domain adaptation problem, we try to promote the domain-invariant feature space. To achieve this, UDA methods add a binary domain discriminator to the output of the backbone \cite{adaptive_teacher,da_wild}. This can be extended to MSDA by having multiple such discriminators. But this doesn't consider the domain discrepancy between the source domain and increases the computational cost \cite{MDAN, multi_disc_class}. Instead of using multiple binary discriminators, we overcome these issues by introducing a multi-class discriminator. This discriminator is connected to the network using a gradient reversal layer (GRL) \cite{grl}. It seeks to classify images according to a domain. When this classification loss is back-propagated, the GRL layer reverses the gradients, so the backbone network will try to increase the classification loss, challenging the ability of the discriminator to distinguish between domains. Gradually, the domain of the image-level features extracted by the backbone network will be indistinguishable from the discriminator. This simple adversarial game between the discriminator and the backbone allows us to learn domain-invariant features \cite{da_wild}. The discriminator is trained with the cross-entropy loss between the $\hat{y}^j$, the $i$-th one-hot vector for the domain $j$, and the output of the domain discriminator on image $x^j_i$: 
\begin{equation}\label{eq:disc}
	\begin{aligned}
	\mathcal{L}_{\mbox{dis}} = - \sum_{j = 1}^{N+1} \sum_{i = 1}^{M_i} \hat{y}^j  \log(D(G(F(x_i^j))) 
	\end{aligned}
\end{equation}
where $F$ is the CNN feature extractor, $G$ is the GRL, and $D$ is the multi-domain discriminator.
The summation goes from $j = 1, 2, ... N+1$ as we consider the target domain.

\subsection{Domain-specific Features with Prototypes}

\begin{figure}[!b]
  \includegraphics[height=3.6cm, width=0.48\textwidth]{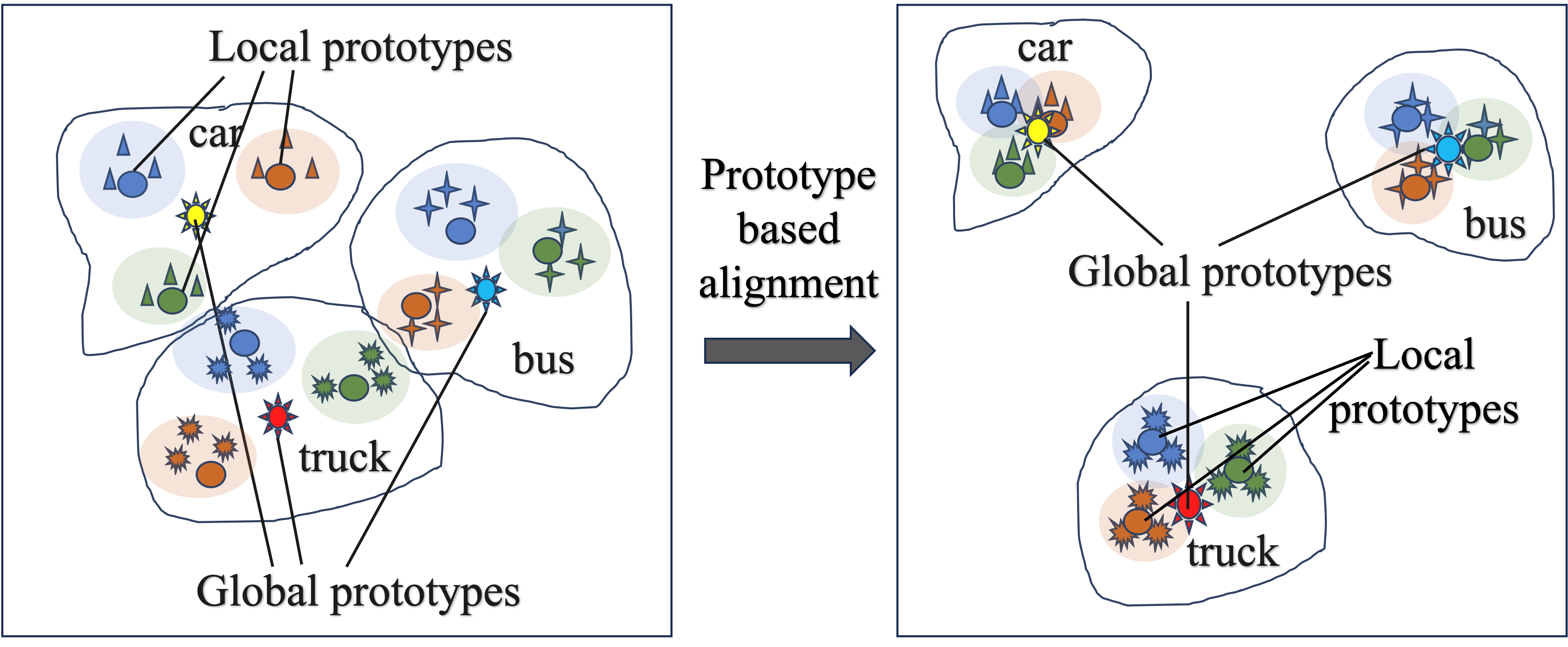}
  \caption{Prototype-based feature alignment with multiple source domains. There are three domains and three classes. Each domain has a prototype for each class. Initially, there is confusion between classes and the intra-class distance to global prototypes from multiple domains is also large. After alignment, class confusion and intra-class distance to global prototypes are reduced.}
  \label{fig:prot_alighment}
\end{figure}

In our PMT method, prototypes are used to preserve domain-specific information. Preserving domain-specific information helps for distilling with confident pseudo labels from individual source domains. To obtain the prototypes, we added a prototype network to the output of the ROI pool features. This network maps the output of the ROI pool to a $d$ dimensional feature space, where class-conditioned feature alignment is performed using a contrastive loss. This aligns representations of the same object categories across different domains and separates representations of different object categories. \cref{fig:prot_alighment} illustrates the prototypes working for domain-specific feature alignment. We produce a local prototype for each class (e.g., car, truck, bus in \cref{fig:prot_alighment}) in each domain, and these prototypes preserve the domain-specific information. The prototype for each class ``across all domains", called global prototypes, is obtained as the mean of the local prototypes of that class. The contrastive loss on global prototypes pushes them apart in the feature space, resulting in better separation of the classes. It also aligns the local prototypes from all domains for each class, reducing confusion due to appearance variation. Note that the prototype network shares parameters across domains, so its parameter size won't change regardless of the number of domains used.
\setlength{\belowcaptionskip}{-5pt}
Let $p_k^j$ denote the local prototype of the $k$th class in domain $j$. The global prototype of class $k$ is denoted as $P_k$. Let the number of occurrences of class $k$ from domain $j$ in a minibatch be $c_k^j$. The prototype update value for class $k$ from domain $j$ in a minibatch is computed as
\begin{equation}
	q_k^j = \frac{1}{|c_k^j|} \sum_{r = 1}^{c_k^j} {P}({R}({F}(x^j), r))
\end{equation}
where $P$ is the prototype network (a 2-layer MLP), $R$ is the faster R-CNN ROI-pooling and $F$ is the CNN feature extractor. Thus, ${R}({F}(x^j), r)$ pools features from the RoI $r$ corresponding to the ground-truth box for class $k$ from domain $j$ in a minibatch. For the target domain without ground-truth annotations, the pseudo-labels obtained from the teacher network are used to compute the update value $q_k^{N+1}$. When $c_k^j=0$ in a minibatch, no updates are performed for the corresponding prototype. The local prototype vector $p_k^j$ for each class $k$ and domain $j$ is stored in memory along with a count value $\rho_k^j$ that tells how many times $p_k^j$ has been updated thus far. Given the current update value $q_k^j$, local prototype $p_k^j$ is updated as:
\begin{equation}
    p_k^j = \frac{\rho_k^j p_k^j + q_k^j}{\rho_k^j+1}
\end{equation}

The count of updates $\rho_k^j$ is incremented after this operation.
After updating the prototype vectors, the contrastive loss is computed in two parts. The first part aligns the prototypes of the same class from different domains by maximizing their similarity. The second part pushes apart the prototypes of different classes to reduce class confusion. To align the prototypes of the same class from different domains, we need to consider pairwise combinations of the domains: $C=Comb(N+1,2)$ (including source and target domains). Let $\mathcal{C}$ be the cosine similarity between two vectors. The alignment of category $k$ across different domains is achieved by maximizing the following similarity:
\begin{equation}
\frac{1}{CK} \sum_{k=1}^K \sum_{j=1}^{N+1} \sum_{l \neq j}^{N+1} \mathcal{C}(p_k^j, p_k^l)
\end{equation}
normalized by the total number of pairwise similarities $CK$. To push the prototypes of the different classes apart, we minimize their cosine similarity. For this, we first compute the global prototypes from local prototypes of each domain as their weighted mean:
\begin{equation}
P_k = \frac{\sum_{j=1}^N \rho_k^j p_k^j} {\sum_{j=1}^N \rho_k^j}
\end{equation}
where the weight $\frac{\rho_k^j}{\sum_{j=1}^N \rho_k^j}$ assigns importance to the prototype of class $k$ from domain $j$ according to its number of occurrence $\rho_k^j$. Once $P_k$ for each class is computed, we minimize the following similarity function:
\begin{equation}
    \sum_{k=1}^K \sum_{l \neq k}^K \mathcal{C}(P_k, P_l) 
\end{equation}
Note that prototypes from the target domain are not considered here because of the noise they induce due to the use of pseudo-labels. The final contrastive loss on prototypes is:
\begin{equation}
\begin{aligned}
    \mathcal{L}_{\mbox{prot}} = \sum_{k=1}^K \sum_{l \neq k}^K \mathcal{C}(P_k, P_l) - 
   \frac{1}{CK} \sum_{k=1}^K \sum_{j=1}^{N+1} \sum_{l \neq j}^{N+1} \mathcal{C}(p_k^j, p_k^l)
\end{aligned}
\label{eq:loss_prot}
\end{equation}

\subsection{Training Summary }

The FRCNN model is initialized with Imagenet pre-trained weights. The weights of the student model are updated using back-propagation. In contrast, the weights of the teacher model are obtained as the EMA of the student model's weights over time. Training starts with a burn-in training stage, where only the Eq. \ref{eq:super} is used to train the student model. Then, the teacher model is initialized as a copy of the student model, and then the weights of the student model are adapted using the back-propagation, while the teacher weights are the EMA of the student model's weights. In the unsupervised MSDA training step, all the above losses are used jointly to adapt the student model. The total loss is: 
\begin{equation}\label{eq:total_loss}
	\begin{aligned}
	\mathcal{L}= \mathcal{L}_{\mbox{sup}} + \alpha \mathcal{L}_{\mbox{unsup}} + \beta \mathcal{L}_{\mbox{dis}} + \gamma \mathcal{L}_{\mbox{prot}}
	\end{aligned}
\end{equation}
where hyperparameters $\alpha$, $\beta$, and $\gamma$ weight the contribution of each loss. For inference, only the teacher model is used to predict output detections.

%% file: 4-experiments.tex
\section{Results and Discussion}
\label{sec:experiments}

\subsection{Experimental Methodology} \label{sec:methodology}

\noindent \textbf{Implementation Details.} 
Our experiments follow the same procedure as in \cite{trkp,dmsn}. FRCNN\cite{fasterrcnn} VGG16 backbone pre-trained with ImageNet was used as the detection framework. Similar to \cite{maskrcnn}, ROI-alignment was used, and the shorter side of the input image was reduced to 600 pixels. For our mean-teacher training, the same setting was used as in \cite{unbiased_teac}. For filtering the pseudo-labels a threshold of 0.7 is used. The model is trained in the burn-in setting for 15 epochs. Then, the model was trained in the unsupervised DA setting for another 15 epochs. For all the settings, the value $\alpha$ and $\beta$ were set to 1 and 0.1, respectively, while the value of $\gamma$ was set to 1.2, 0.6, and 0.1 for Sections \ref{sec:cross_time}, \ref{sec:cross_camera} and \ref{sec:mixed_da}. The weight smoothing coefficient of EMA is set to 0.9996. We conducted all the experiments on 4 A100 GPUs, with batch size 4 and a learning rate of 0.02. Our approach was implemented using Pytorch\cite{pytorch} and Detectron2\cite{detectron2}. 

\noindent \textbf{Quantitative comparison.} We compared our approach with the following baseline: (1) Source-Only: FRCNN\cite{fasterrcnn} is trained in a supervised manner on source domain data, and no adaptation is performed. (2) UDA Blending: all the source domain data is combined, and UDA methods \cite{strongweak, crossdomain_proto, categorical, unbiased_teac, adaptive_teacher} are used to adapt FRCNN. (3) MSDA: adaptation of FRCNN from independent source datasets. It includes MSDA methods developed for classification \cite{moment,MDAN} and upgraded for OD, as well as MSDA methods developed for OD \cite{dmsn}\cite{trkp}. (4) Oracle: Target-Only and All-Combined. In the Target-Only case, we performed supervised training of FRCNN only on labeled target data. In the All-Combined case, supervised training is performed on all source and target data combined. Note that some of the baseline results are taken from \cite{dmsn} and \cite{trkp}.
\subsection{Cross Time Adaptation} 
\label{sec:cross_time}
In this setting, images are collected at different times of the day, so domain shift is caused by changes in illumination. The performance of our method in this setting is evaluated on the BDD100K\cite{bdd100k} dataset. The dataset consists of data collected at three different times - Daytime, Dusk/Dawn, and Night Time. In our experiments, Daytime and Night are used as the source domains, and Dusk/Dawn as the target domain. The source domain consists of labeled 64,699 (36,728 daytime + 27,971 night) images. In the target domain, we have 5,027 unlabeled images, that are used for training. The evaluation is done on 778 validation set images of Dusk/Dawn. The mAP on 10 classes is reported. 

The mAP performance of our method is compared against the other approaches in Table~\ref{table:BDD100k_sota}. Class-wise AP is reported in Table~\ref{table:Classwise_bdd100k} in Appendix. It can be observed that the UDA methods that blend source datasets are able to increase the performance of the detector compared to the Source-Only baseline. However, this increase in performance is not large, as they disregard the inter-source domain shift. It is also interesting to note that two state-of-the-art MSDA classification methods \cite{MDAN, moment}, perform worse than the Source-Only baseline. This degradation in performance shows the fundamental difference in MSDA methods for classification and detection and a strong motivation for MSDA for OD. Our method improves on the performance of the best-performing MSDA method \cite{trkp} by 5.5\%. It can also be seen that the performance of our PMT method is much better than the Oracle Target-Only case (because of the small size of the target dataset), and it approached the mAP of the Oracle All-Combined case. 
\begin{table}[!t]
    \centering
    \resizebox{.49\textwidth}{!}{
      \begin{tabular}{c l | c c c}
      \hline
    \textbf{Setting} & \textbf{Method} & \textbf{D} & \textbf{N} & \textbf{D+N} \\ 
    \hline
    \hline
    Source Only & FRCNN\cite{fasterrcnn} & 30.4 & 25.0 & 28.9\\
     \hline
     
    \multirow{5}{*}{UDA Blending} & Strong-Weak\cite{strongweak} & 31.4 & 26.9 & 29.9\\
    & Graph Prototype\cite{MT_graph} & 31.8 & 27.6 & 30.6\\
    & Cat. Regularization\cite{categorical} & 31.2 & 28.4 & 30.2\\
    & UMT\cite{MTunbiased} & 33.8 & 21.6 & 33.5 \\
    & Adaptive Teacher\cite{adaptive_teacher} & 34.9 & 27.8 & 34.6 \\
     
    \hline

    \multirow{4}{*}{MSDA} & MDAN\cite{MDAN} & - & - & 27.6 \\
    & M$^3$SDA\cite{moment} & - & - & 26.5 \\
    & DMSN\cite{dmsn} & - & -& 35.0 \\
    & TRKP\cite{trkp} & - & - & 39.8 \\
    & \textbf{PMT(ours)} & - & - & \textbf{45.3} \\
    \hline

    \multirow{2}{*}{Oracle} & Target-Only & - & - & 26.6 \\
    &  All-Combined & - & - & 45.6 \\
    \hline
    \end{tabular}
    }
    \caption{Detection AP of PMT compared against the baseline, UDA, MSDA, and oracle methods on BDD100K. Source domains are daytime (D) and night (N) subsets and the target is always Dusk/Dawn of BDD100K. 
    }
    \label{table:BDD100k_sota}
\end{table}

\subsection{Cross Camera Adaptation} \label{sec:cross_camera}
In this setting, the data are collected using different cameras, and domain shift is caused by changes in the camera's resolution and viewpoint. In this setting BDD100K\cite{bdd100k}, Cityscape\cite{cityscapes}, and Kitty\cite{kitty} datasets are used. For our experiments, we used Cityscape and Kitty as the source domains, while the Daytime domain of BDD100k was used as the target domain. For both training and evaluation, we only considered the images with car objects. This provides a source domain that consists of 9,515 (2,831 Cityscapes + 6,684 Kitty) labeled images. In the target domain, we have 36,728 unlabeled images used for training. The evaluation is done on 5,258 validation set images of Daytime. The AP is reported only on the car object category in Table \ref{table:kitty}.

Note that in this experiment, there is only one object category (car), so we cannot use both components of our contrastive loss \ref{eq:loss_prot}. For this case, the prototype separation part was removed for our contrastive loss. It can be observed that the variation in the performance is similar to Section \ref{sec:cross_time}. Our PMT can outperform state-of-the-art methods, but the increase in AP is lower compared to TRKP. We suppose this is due to the removal of the prototype separation part of our contrastive loss. In ablation studies, we showed that every loss component contributed to obtaining the best PMT performance. 

\begin{table}[!t]
    \centering
    \resizebox{.49\textwidth}{!}{%
    \begin{tabular}{c l | c c c}
    \hline
    \textbf{Setting} & \textbf{Method} & \textbf{C} & \textbf{K} & \textbf{C+K} \\ 
    \hline
    \hline
    Source Only & FRCNN\cite{fasterrcnn} & 44.6 & 28.6 & 43.2\\
    
     \hline
    \multirow{4}{*}{UDA Blending} & Strong-Weak\cite{strongweak} & 45.5 & 29.6 & 41.9\\
    & Cat. Regularization\cite{categorical} & 46.5 & 30.8 & 43.6\\
    & UMT\cite{MTunbiased} & 47.5 & 35.4 & 47.0 \\
    & Adaptive Teacher\cite{adaptive_teacher} & 49.8 & 40.1 & 48.4 \\
    \hline
    \multirow{4}{*}{MSDA} & MDAN\cite{MDAN} & - & - & 43.2 \\
    & M$^3$SDA\cite{moment} & - & - & 44.1 \\
    & DMSN\cite{dmsn} & - & -& 49.2 \\
    & TRKP\cite{trkp} & - & - & 58.4 \\
    & \textbf{PMT(ours)} & - & - & \textbf{58.7} \\
    \hline
    \multirow{2}{*}{Oracle} & Target-Only   & - & - & 60.2 \\
                            & All-Combined  & - & - & 69.7 \\
    \hline
    \end{tabular} %
    } 
    \caption{AP for the car class. Our proposed method PMT is compared against the baseline, UDA, MSDA, and oracle methods on the \textit{Daytime} domain of BDD100K dataset. C and K refer to Cityscapes and Kitty datasets.}
    \label{table:kitty}
\end{table}

\subsection{Extension to Mixed Domain Adaptation} 
\label{sec:mixed_da}
In the MSDA, the domain shift among source data is not always limited to one factor. So, a setting with domain shift with multiple factors is considered for validation. In the source domain, we considered the MS COCO\cite{mscoco-Lin-2014}, Cityscapes\cite{cityscapes}, and Synscapes\cite{synscapes} datasets, and the Daytime domain of the BDD100K dataset are the target domain. Among the source domains, the domain shift is mixed and there are more classes, making this study a challenging scenario. Also, the number of source domains is increased from two to three. The source domain has 99,724 (2,975 Cityscapes + 71,749 MS COCO + 25,000 Synscapes) labeled images and, the target domain has 36,728 unlabeled images. The evaluation is performed on 5,258 validation set images of Daytime. For training and evaluation, we only employ the images having 7 common object categories between the datasets. From the results reported in Table \ref{table:coco_sota}, we can observe that the Source-Only setting performs better than the UDA Blending setting. This is because of the complex domain shift among the source data. When only Cityscapes and MS COCO are used as the source domain, our method outperforms the previous best-performing method by 3.4\%. After adding the Synscapes dataset our method still outperforms other methods, including the Oracle Target-Only results. This experiment further highlights the effectiveness of our method. Class-wise AP is reported in Table~\ref{table:Classwise_mixed} in Appendix.

\begin{table}[!t]
    \centering
    \resizebox{.48\textwidth}{!}{
      \begin{tabular}{c l | c c c}
      \hline
    \textbf{Setting} & \textbf{Method} & \textbf{C} & \textbf{C + M} & \textbf{C + M + S} \\ 
    \hline \hline
    Source Only & FRCNN\cite{fasterrcnn} & 23.4 & 29.7 & 30.9\\
     \hline  
   \multirow{2}{*}{UDA Blending} & Unbiased Teach.\cite{unbiased_teac} & - & 18.5 & 25.1\\
    & Adaptive Teach.\cite{adaptive_teacher} & - & 22.9 & 29.6\\
    \hline
    \multirow{2}{*}{MSDA} & TRKP\cite{trkp} & - & 35.3 & 37.1 \\
    & \textbf{PMT(ours)} & - & \textbf{38.7} & \textbf{39.7} \\
     \hline
   \multirow{2}{*}{Oracle} & Target-Only & - & - & 38.6 \\
   & All-Combined & - & 47.1 & 48.2 \\
   \hline
    \end{tabular}
    }
    \caption{ mAP on 7 object categories of our PMT compared against baselines on the \textit{Daytime} domain of BDD100K. C, M, and S refer to Cityscapes, MS COCO, and Synscapes datasets.}
    \label{table:coco_sota}
\end{table}

\subsection{Ablation Studies}

\noindent \textbf{Impact of the different components.} 
The cross-time adaptation setting is used in the ablation studies. The two main components of our method are the prototype network and multi-class discriminator. To clearly disentangle the effect of terms in the contrastive loss for prototypes, we considered them as two separate parts: ``prototype separation'' and ``prototype alignment" part. The prototype separation loss increases the distance between different object categories across domains, whereas the prototype alignment loss aligns the same object category across various domains. The experimental results shown in Table \ref{table:components} show that the results without using the discriminator and prototype are better than the Source-Only result of Table \ref{table:BDD100k_sota}. This is due to the effectiveness of the mean-teacher framework in MSDA. By adding the discriminator to the network, performance is increased by 10.2\%. This result is already better than the previous state of art MSDA methods for OD.  By adding both of our prototype loss terms individually, the performance of the model further increases. Note that using the alignment loss performs better than the separation loss because the separation loss only discriminates the categories, and the vanilla FRCNN is already doing it. Finally, when both the prototype loss terms are combined, the performance is further improved. 

\begin{table}[!t]
    \centering
    \resizebox{.4\textwidth}{!}{
      \begin{tabular}{ccc|c}
    \hline
    \textbf{Multidomain} &   \textbf{Prototype} & \textbf{Prototype} &  \\
    \textbf{Discriminator} &   \textbf{Separation} & \textbf{Alignment} & \textbf{$\textrm{AP}_{50}$} \\
    \hline \hline
     &  &  & 31.7 \\
    $\checkmark$    &               &               & 41.9 \\
    $\checkmark$    & $\checkmark$  &               & 43.0 \\
    $\checkmark$    &               & $\checkmark$  & 43.4\\
    $\checkmark$    & $\checkmark$  &  $\checkmark$ & 44.6\\
    \hline
    \end{tabular}
    }
    \caption{Ablation study on the components of our PMT method.}
    \label{table:components}
\end{table}

\subsection{Parameter Growth with Number of Domains}
One of the important advantages of not using domain-specific weights is the reduction in the number of parameters to learn. While the existing MSDA detectors have a significant linear growth in parameters with the number of domains, the parameter growth of our method is negligible as the number of source domains increases. \cref{table:param_growth} illustrates this. DMSN \cite{dmsn} makes some part of the feature extractor also domain-specific, thus showing a significantly higher rate of growth compared to other methods. TRKP \cite{trkp} makes only the detection heads domain-specific, so the growth rate is less compared to DMSN. Our method has only domain-specific prototypes which require fewer parameters per domain than the other approaches. 

\begin{table}[!t]
    \centering
    \resizebox{.47\textwidth}{!}{
      \begin{tabular}{c | c c c c c}
      \hline
    \textbf{\multirow{2}{*}{Method}} &\multicolumn{5}{c}{Number of source domains}\\
    \cline{2-6}
    & 1 & 2 & 3 & 4 & 5\\
    \hline \hline
   DMSN \cite{dmsn} & 45.994 & 75.426 & 104.858 & 134.290 & 163.722 \\
   TRKP \cite{trkp} & 45.994 & 59.942 & 73.890 & 87.838 & 101.786 \\
   \textbf{PMT (ours)} & 46.586 & 46.587 & 46.588 & 46.589 & 46.590\\
   \hline
    \end{tabular}
    }
    \caption{Model parameter growth (in millions) as the number of source domains increases. While the parameters of DMSN \cite{dmsn} and TRKP \cite{trkp} grow quickly with the number of source domains, the parameters of our method PMT remain almost constant because the increase is only due to the prototype vectors.}
    \label{table:param_growth}
\end{table}

%% file: 5-conclusion.tex
\section{Conclusion}
\label{sec:conclusion}
A commonly used architecture for MSDA is to learn domain-invariant and domain-specific parameters (for each source domain) to effectively adapt DL models from multiple source domains. In this paper, we proposed a method for simplifying MSDA where domain-specific information is retained using class prototypes for each domain. This avoids the need to train a domain-specific subnet for each source domain, simplifying the MSDA architecture significantly. The parameters of our model remain almost constant regardless of the number of source domains used as only prototype vectors are needed for each domain. Thus, our model size is comparable to UDA methods which perform adaptation from a single source. Experimental results show that our proposed method can effectively exploit multiple domains and improve on the state-of-the-art for multi-source object detection.

%% file: 6-supplementary.tex
\section{Appendix}

\subsection{Impact of $\gamma$ on different settings} 
The main component responsible for simplifying the MSDA problem with our approach is prototype-based learning. 
In this study, we investigate the importance of contrastive loss on prototypes by studying the impact of $\gamma$ in different settings. For experiments, we considered the two cases; Cross Time Adaptation and Mixed Domain Adaptation settings. The Cross Time Adaptation setting is much simpler compared to the Mixed Domain Adaptation setting. By choosing these settings, we get an idea of the effect of $\gamma$ on simple and complex MSDA problems. The analysis result is reported in Table \ref{table:gamma}. 
It can be observed that for the Cross Time Adaptation setting, the performance of our method increased with the increase in $\gamma$, while it decreased for the Mixed Domain Adaptation setting. With the increase in value of $\gamma$, the model is forced to learn a class-conditioned aligned feature space. Learning this feature space is much easier if the domain shift among the datasets is not much. But with the increase in domain shift, learning the aligned feature space becomes difficult. 

\begin{table}[h]
    \centering
    \resizebox{.8\linewidth}{!}{
      \begin{tabular}{c | c | c | c }
      \hline
    \textbf{$\gamma$} & \textbf{Source Domain} & \textbf{Target Domain} & \textbf{mAP}\\ 
    \hline
    \hline
     0.1 & \multirow{5}{*}{D+N} & \multirow{5}{*}{Dusk/Dawn} & 43.6\\
     0.5 &  &  & 43.9\\
     0.9 &  &  & 44.6\\
     1.2 &  &  & 45.3\\
     1.5 &  &  & 45.1\\
     \hline
     0.1 & \multirow{5}{*}{C+M+S} & \multirow{5}{*}{Daytime} & 39.7\\
     0.5 &  &  & 39.1\\
     0.9 &  &  & 38.7\\
     1.2 &  &  & 38.3\\
     1.5 &  &  & 38.2\\
     \hline
    \end{tabular}
    }
    \caption{Effect of $\gamma$ under different domain shift conditions.}
    \label{table:gamma}
\end{table}

\subsection{Impact of the Number of Classes} 
\begin{figure}[!b]
  \centering
  \includegraphics[ width=0.4\textwidth]{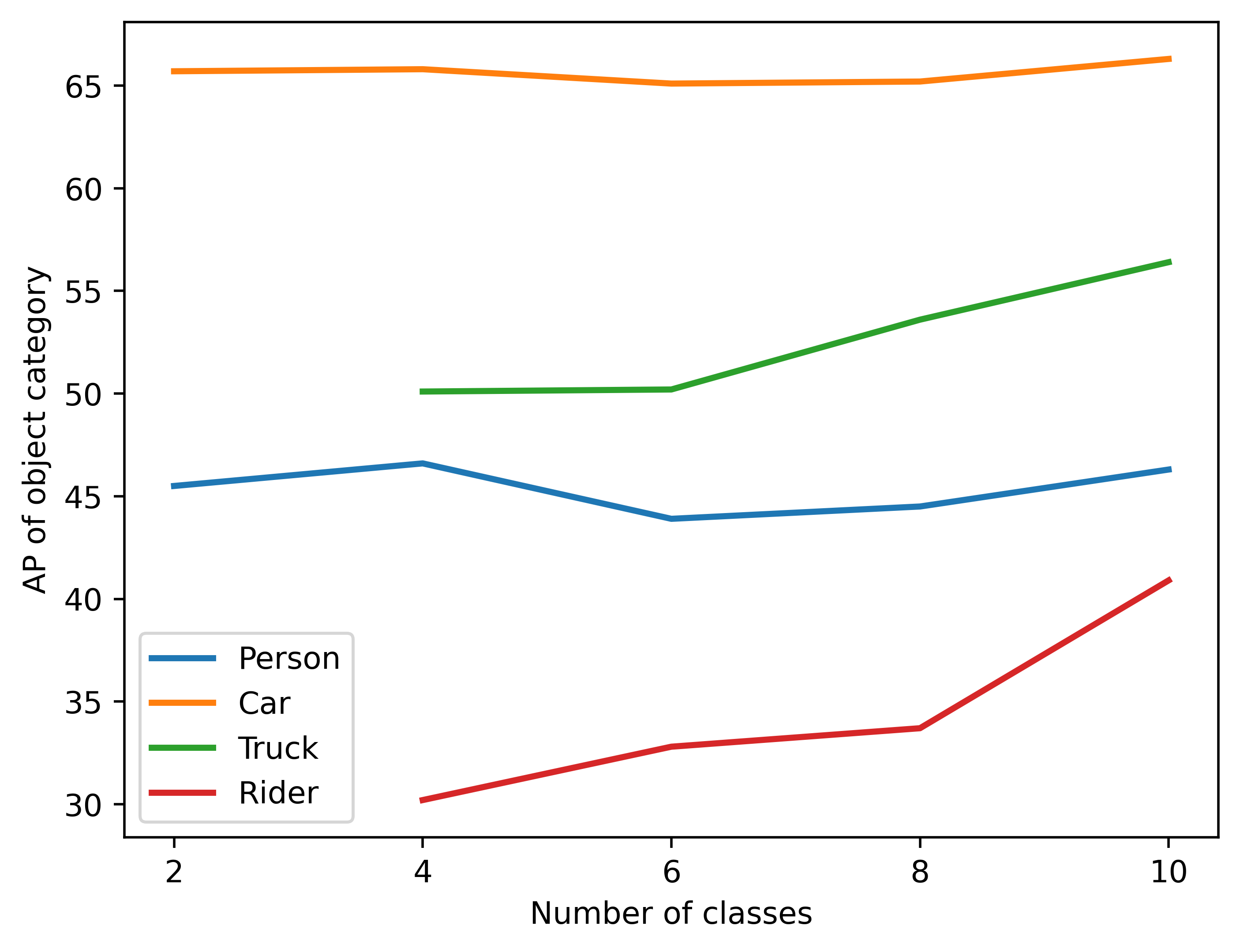}
  \caption{AP of PMT with a growing number of object categories.}
  \label{fig:classes}
\end{figure}

In this ablation, we consider the effect of a different number of object categories for training and evaluation on our PMT. We analyze the performance of a given class when trained with a varying number of other classes and how this affects the final detection performance. As the prototypes are trained with a contrastive loss, we expect that an increased number of classes could help to further improve the overall adaptation.
In \cref{fig:classes}, we present the AP for the classes Person and Car when trained with 0 to 8 additional classes and Truck and Raider when trained with 2 to 8 extra classes. While for some of the classes (Person and Car), the effect is not clear, for others (Truck and Rider) there is a clear trend in which more classes are used and better the results. We hypothesize that this different behavior could be due to the number of training samples per class. For very common classes with many samples, the model is already quite general, and adding more classes does not help. Instead for classes with fewer data points, contrasting the prototypes with other classes helps to improve detection. 

\begin{table*}
    \centering
    \resizebox{.90\textwidth}{!}{
      \begin{tabular}{l l l | l l l l l l l l l l | l }
      \hline
    \textbf{Setting} & \textbf{Source} & \textbf{Method}  & \textbf{Bike} & \textbf{Bus} & \textbf{Car} &
    \textbf{Motor} & \textbf{Person} &
    \textbf{Rider} & \textbf{Light} & \textbf{Sign} & \textbf{Train} & \textbf{Truck} & \textbf{mAP}\\ 
    \hline
    \hline
     \multirow{3}{*}{Source Only} & D & \multirow{3}{*}{FRCNN\cite{fasterrcnn}} & 35.1 & 51.7 & 52.6 & 9.9 & 31.9 & 17.8 & 21.6 & 36.3 & - & 47.1 & 30.4
     \\
     & N &  & 27.9 & 32.5 & 49.4 & 15.0 & 28.7 & 21.8 & 14.0 & 30.5 & - & 30.7 & 25.0
     \\
     & D+N &  & 31.5 & 46.9 & 52.9 & 8.4 & 29.5 & 21.6 & 21.7 & 34.3 & - & 42.2 & 28.9
     \\
     \hline

     \multirow{5}{*}{UDA Blending} & \multirow{5}{*}{D+N} & Strong-Weak\cite{strongweak} & 29.7 & 50.0 & 52.9 & 11.0 & 31.4 & 21.1 & 23.3 & 35.1 & - & 44.9 & 29.9
     \\
     &  & Graph Prototype\cite{MT_graph} & 31.7 & 48.8 & 53.9 & 20.8 & 32.0 & 21.6 & 20.5 & 33.7 & - & 43.1 & 30.6
     \\
     &  & Cat. Regularization\cite{categorical} & 25.3 & 51.3 & 52.1 & 17.0 & 33.4 & 18.9 & 20.7 & 34.8 & - & 47.9 & 30.2
     \\
     &  & UMT\cite{MTunbiased} & 42.3 & 48.1 & 56.4 & 13.5 & 35.3 & 26.9 & 31.1 & 41.7 & - & 40.1 & 33.5
     \\
     &  & Adaptive Teacher\cite{adaptive_teacher} & 43.1 & 48.9 & 56.9 & 14.7 & 36.0 & 27.1 & 32.7 & 43.8 & - & 42.7 & 34.6
     \\
     \hline

     \multirow{5}{*}{MSDA} & \multirow{5}{*}{D+N} & MDAN\cite{MDAN} &  37.1 & 29.9 & 52.8 & 15.8 & 35.1 & 21.6 & 24.7 & 38.8 & - & 20.1 & 27.6
     \\
     & & M$^3$SDA\cite{moment} & 36.9 & 25.9 & 51.9 & 15.1 & 35.7 & 20.5 & 24.7 & 38.1 & - & 15.9 & 26.5
     \\
     & & DMSN\cite{dmsn} & 36.5 & 54.3 & 55.5 & 20.4 & 36.9 & 27.7 & 26.4 & 41.6 & - & 50.8 & 35.0
     \\
     & & TRKP\cite{trkp} & 48.4 & 56.3 & 61.4 & 22.5 & 41.5 & 27.0 & 41.1 & 47.9 & - & 51.9 & 39.8
     \\
     &  & \textbf{PMT(ours)} & \textbf{55.3}  & \textbf{59.8}  & \textbf{67.6}  &  \textbf{29.9}  &  \textbf{47.6} & \textbf{32.7}  &  \textbf{46.3}   & \textbf{56.0}  &  -  &  \textbf{57.7} & \textbf{45.3}
     \\

    \hline
     \multirow{2}{*}{Oracle} & \multirow{2}{*}{D+N} & Target Only & 27.2  & 39.6  & 51.9   &  12.7  &  29.0 & 15.2  &  20.0   & 33.1  &  -  &  37.5 & 26.6
     \\
      & &  All-Combined & 56.4  & 59.9  & 67.3  &  30.8  &  47.9 & 33.9  &  47.2   & 57.8  &  -  &  54.8 & 45.3
      \\
     \hline
    \end{tabular}
    }
    \caption{Class-wise AP of PMT compared against the baseline,
UDA, MSDA, and oracle methods on BDD100K. Source domains
are daytime (D) and night (N) subsets and the target is always
Dusk/Dawn of BDD100K}
    \label{table:Classwise_bdd100k}
\end{table*}

\begin{table*}
    \centering
    \resizebox{.90\textwidth}{!}{
      \begin{tabular}{l l l | l l l l l l l  | l }
      \hline
    \textbf{Setting} & \textbf{Source} & \textbf{Method}  & \textbf{Person} & \textbf{Car}  &
    \textbf{Rider} & \textbf{Truck} &
    \textbf{Motor} & \textbf{Bicycle} & \textbf{Bike} & \textbf{mAP}\\ 
    \hline
    \hline
     Source Only & C & FRCNN\cite{fasterrcnn}  & 26.9 & 44.7 & 22.1 & 17.4 & 17.1 & 18.8 & 16.7 & 23.4
     \\
    \hline
     Source Only &\multirow{5}{*} {C+M} & FRCNN\cite{fasterrcnn} & 35.2 & 49.5 & 26.1 & 25.8 & 18.9 & 26.1 & 26.5 & 29.7
     \\
    UDA Blending &  & Unbiased Teach.\cite{unbiased_teac} & 30.7 & 28.0 & 3.9 & 11.2 & 19.2 & 17.8 & 18.7 & 18.5
    \\
    UDA Blending &  & Adaptive Teacher\cite{adaptive_teacher} & 31.2 & 31.7 & 15.1 & 16.4 & 17.1 & 20.9 & 27.9 & 22.9
    \\
    MSDA &  &  TRKP\cite{trkp} & 39.2 & 53.2 & \textbf{32.4} & 28.7 & 25.5 & 31.1 & 37.4 & 35.3
     \\
     MSDA &  & \textbf{PMT(ours)} & \textbf{41.1} & \textbf{53.5} & 31.2 & \textbf{31.9} & \textbf{33.7} & \textbf{34.9} & \textbf{44.6} & \textbf{38.7}
     \\
    \hline
     
     Source Only & \multirow{5}{*}{C+M+S} & FRCNN\cite{fasterrcnn} & 36.6 & 49.0 & 22.8 & 24.9 & 26.9 & 28.4 & 27.7 & 30.9
     \\
      UDA Blending &  & Unbiased Teach.\cite{unbiased_teac} & 32.7 & 39.6 & 6.6 & 21.2 & 21.3 & 25.7 & 28.5 & 25.1
     \\
     UDA Blending &  & Adaptive Teacher\cite{adaptive_teacher} & 36.3 & 42.6 & 19.7 & 23.4 & 24.8 & 27.1 & 33.2 & 29.6
     \\
      MSDA &  &  TRKP\cite{trkp} & 40.2 & 53.9 & 31.0 & 30.8 & 30.4 & 34.0 & 39.3 & 37.1
     \\
     MSDA &  & \textbf{PMT(ours)} & \textbf{43.3} & \textbf{54.1} & \textbf{32.0} & \textbf{32.6} & \textbf{35.1} & \textbf{36.1} & \textbf{44.8} & \textbf{39.7}
     \\
     \hline

     \multirow{3}{*}{Oracle} & C+M &  Target Only & 35.3 & 53.9 & 33.2 & 46.3 & 25.6 & 29.3 & 46.7 & 38.6
     \\
     & C+M & All-Combined & 40.2 & 60.1 & 47.1 & 60.0 & 29.2 & 36.3 & 56.9 & 47.1
     \\
       & C+M+S & All-Combined & 41.7 & 63.9 & 49.5 & 58.1 & 31.6 & 39.1 & 53.5 & 48.2
     \\
     \hline
    \end{tabular}
    }
    \caption{Class-wise AP of PMT compared against the baselines on Daytime domain of BDD100K. C, M, and S refer to Cityscapes, MS COCO, and Synscapes datasets.}
    \label{table:Classwise_mixed}
\end{table*}

%% file: PaperForReview.bbl
\begin{thebibliography}{10}\itemsep=-1pt

\bibitem{MT_graph}
Qi Cai, Yingwei Pan, Chong-Wah Ngo, Xinmei Tian, Lingyu Duan, and Ting Yao.
\newblock Exploring object relation in mean teacher for cross-domain detection, 2019.

\bibitem{detr-Carion-2020}
N. Carion, F. Massa, G. Synnaeve, N. Usunier, A. Kirillov, and S. Zagoruyko.
\newblock End-to-end object detection with transformers.
\newblock In {\em ECCV}, 2020.

\bibitem{da_wild}
Yuhua Chen, Wen Li, Christos Sakaridis, Dengxin Dai, and Luc~Van Gool.
\newblock Domain adaptive faster r-cnn for object detection in the wild, 2018.

\bibitem{cityscapes}
Marius Cordts, Mohamed Omran, Sebastian Ramos, Timo Rehfeld, Markus Enzweiler, Rodrigo Benenson, Uwe Franke, Stefan Roth, and Bernt Schiele.
\newblock The cityscapes dataset for semantic urban scene understanding.
\newblock In {\em Proc. of the IEEE Conference on Computer Vision and Pattern Recognition (CVPR)}, 2016.

\bibitem{MTunbiased}
Jinhong Deng, Wen Li, Yuhua Chen, and Lixin Duan.
\newblock Unbiased mean teacher for cross-domain object detection, 2021.

\bibitem{voc-Everingham-2010}
M. Everingham, L.~Van Gool, C.~K. Williams, J. Winn, and A. Zisserman.
\newblock The pascal visual object classes (voc) challenge.
\newblock {\em IJCV}, 2010.

\bibitem{grl}
Yaroslav Ganin and Victor Lempitsky.
\newblock Unsupervised domain adaptation by backpropagation, 2015.

\bibitem{kitty}
Andreas Geiger, Philip Lenz, and Raquel Urtasun.
\newblock Are we ready for autonomous driving? the kitti vision benchmark suite.
\newblock In {\em Conference on Computer Vision and Pattern Recognition (CVPR)}, 2012.

\bibitem{maskrcnn}
Kaiming He, Georgia Gkioxari, Piotr Dollár, and Ross Girshick.
\newblock Mask r-cnn, 2018.

\bibitem{progressive}
Han-Kai Hsu, Chun-Han Yao, Yi-Hsuan Tsai, Wei-Chih Hung, Hung-Yu Tseng, Maneesh Singh, and Ming-Hsuan Yang.
\newblock Progressive domain adaptation for object detection, 2019.

\bibitem{openWOD-Joseph-2021}
K~J Joseph, Salman Khan, Fahad~Shahbaz Khan, and Vineeth~N Balasubramanian.
\newblock Towards open world object detection.
\newblock In {\em Proceedings of the IEEE/CVF Conference on Computer Vision and Pattern Recognition (CVPR 2021)}, 2021.

\bibitem{pseudo_classifier}
Mehran Khodabandeh, Arash Vahdat, Mani Ranjbar, and William~G. Macready.
\newblock A robust learning approach to domain adaptive object detection, 2019.

\bibitem{pseudo_ssd}
Seunghyeon Kim, Jaehoon Choi, Taekyung Kim, and Changick Kim.
\newblock Self-training and adversarial background regularization for unsupervised domain adaptive one-stage object detection.
\newblock In {\em 2019 IEEE/CVF International Conference on Computer Vision (ICCV)}, pages 6091--6100, 2019.

\bibitem{ssod_Prot-Li-2022}
Aoxue Li, Peng Yuan, and Zhenguo Li.
\newblock Semi-supervised object detection via multi-instance alignment with global class prototypes.
\newblock In {\em 2022 IEEE/CVF Conference on Computer Vision and Pattern Recognition (CVPR)}, 2022.

\bibitem{adaptive_teacher}
Yu-Jhe Li, Xiaoliang Dai, Chih-Yao Ma, Yen-Cheng Liu, Kan Chen, Bichen Wu, Zijian He, Kris Kitani, and Peter Vajda.
\newblock Cross-domain adaptive teacher for object detection.
\newblock In {\em CVPR}, 2022.

\bibitem{mscoco-Lin-2014}
T.-Y. Lin, M. Maire, S. Belongie, J. Hays, P. Perona, D. Ramanan, P. Dollar, and C.~L. Zitnick.
\newblock Microsoft {COCO}: Common objects in context.
\newblock In {\em ECCV}, 2014.

\bibitem{ssd}
Wei Liu, Dragomir Anguelov, Dumitru Erhan, Christian Szegedy, Scott Reed, Cheng-Yang Fu, and Alexander~C. Berg.
\newblock {SSD}: Single shot {MultiBox} detector.
\newblock In {\em Computer Vision {\textendash} {ECCV} 2016}, pages 21--37. Springer International Publishing, 2016.

\bibitem{unbiased_teac}
Yen-Cheng Liu, Chih-Yao Ma, Zijian He, Chia-Wen Kuo, Kan Chen, Peizhao Zhang, Bichen Wu, Zsolt Kira, and Peter Vajda.
\newblock Unbiased teacher for semi-supervised object detection, 2021.

\bibitem{multi_disc_class}
G. Park and S.~Wan Lee.
\newblock Information-theoretic regularization for multi-source domain adaptation.
\newblock In {\em 2021 IEEE/CVF International Conference on Computer Vision (ICCV)}, pages 9194--9203, 2021.

\bibitem{pytorch}
Adam Paszke, Sam Gross, Francisco Massa, Adam Lerer, James Bradbury, Gregory Chanan, Trevor Killeen, Zeming Lin, Natalia Gimelshein, Luca Antiga, Alban Desmaison, Andreas Kopf, Edward Yang, Zachary DeVito, Martin Raison, Alykhan Tejani, Sasank Chilamkurthy, Benoit Steiner, Lu Fang, Junjie Bai, and Soumith Chintala.
\newblock Pytorch: An imperative style, high-performance deep learning library.
\newblock In {\em Advances in Neural Information Processing Systems 32}, pages 8024--8035. Curran Associates, Inc., 2019.

\bibitem{moment}
Xingchao Peng, Qinxun Bai, Xide Xia, Zijun Huang, Kate Saenko, and Bo Wang.
\newblock Moment matching for multi-source domain adaptation, 2019.

\bibitem{yolo}
Joseph Redmon, Santosh Divvala, Ross Girshick, and Ali Farhadi.
\newblock You only look once: Unified, real-time object detection, 2016.

\bibitem{fasterrcnn}
S. Ren, K. He, R. Girshick, and J. Sun.
\newblock Faster {R-CNN}: Towards real-time object detection with region proposal networks.
\newblock In {\em NeurIPS}, 2015.

\bibitem{strongweak}
Kuniaki Saito, Yoshitaka Ushiku, Tatsuya Harada, and Kate Saenko.
\newblock Strong-weak distribution alignment for adaptive object detection, 2019.

\bibitem{mean_teacher-Antti-2018}
Antti Tarvainen and Harri Valpola.
\newblock Mean teachers are better role models: Weight-averaged consistency targets improve semi-supervised deep learning results.
\newblock In {\em NeurIPS}, 2018.

\bibitem{fcos}
Zhi Tian, Chunhua Shen, Hao Chen, and Tong He.
\newblock Fcos: Fully convolutional one-stage object detection, 2019.

\bibitem{synscapes}
Magnus Wrenninge and Jonas Unger.
\newblock Synscapes: A photorealistic synthetic dataset for street scene parsing, 2018.

\bibitem{fewshot_Prot-Wu-2021}
Aming Wu, Yahong Han, Linchao Zhu, Yi Yang, and Cheng Deng.
\newblock Universal-prototype augmentation for few-shot object detection.
\newblock 2021.

\bibitem{trkp}
Jiaxi Wu, Jiaxin Chen, Mengzhe He, Yiru Wang, Bo Li, Bingqi Ma, Weihao Gan, Wei Wu, Yali Wang, and Di Huang.
\newblock Target-relevant knowledge preservation for multi-source domain adaptive object detection, 2022.

\bibitem{detectron2}
Yuxin Wu, Alexander Kirillov, Francisco Massa, Wan-Yen Lo, and Ross Girshick.
\newblock Detectron2.
\newblock \url{https://github.com/facebookresearch/detectron2}, 2019.

\bibitem{categorical}
Chang-Dong Xu, Xing-Ran Zhao, Xin Jin, and Xiu-Shen Wei.
\newblock Exploring categorical regularization for domain adaptive object detection, 2020.

\bibitem{msda_prototype_cls-Xu-2022}
Minghao Xu, Hang Wang, and Bingbing Ni.
\newblock Graphical modeling for multi-source domain adaptation.
\newblock {\em IEEE Transactions on Pattern Analysis and Machine Intelligence}, 2022.

\bibitem{da_graph-Xu-2020}
Minghao Xu, Hang Wang, Bingbing Ni, Qi Tian, and Wenjun Zhang.
\newblock Cross-domain detection via graph-induced prototype alignment.
\newblock In {\em CVPR}, pages 12355--12364, 2020.

\bibitem{crossdomain_proto}
Minghao Xu, Hang Wang, Bingbing Ni, Qi Tian, and Wenjun Zhang.
\newblock Cross-domain detection via graph-induced prototype alignment, 2020.

\bibitem{dmsn}
Xingxu Yao, Sicheng Zhao, Pengfei Xu, and Jufeng Yang.
\newblock Multi-source domain adaptation for object detection, 2021.

\bibitem{bdd100k}
Fisher Yu, Haofeng Chen, Xin Wang, Wenqi Xian, Yingying Chen, Fangchen Liu, Vashisht Madhavan, and Trevor Darrell.
\newblock Bdd100k: A diverse driving dataset for heterogeneous multitask learning, 2020.

\bibitem{da_Prot-Zhang-2021}
Yixin Zhang, Zilei Wang, and Yushi Mao.
\newblock Rpn prototype alignment for domain adaptive object detector.
\newblock In {\em 2021 IEEE/CVF Conference on Computer Vision and Pattern Recognition (CVPR)}, 2021.

\bibitem{MDAN}
Han Zhao, Shanghang Zhang, Guanhang Wu, Jos\'{e} M.~F. Moura, Joao~P Costeira, and Geoffrey~J Gordon.
\newblock Adversarial multiple source domain adaptation.
\newblock In S. Bengio, H. Wallach, H. Larochelle, K. Grauman, N. Cesa-Bianchi, and R. Garnett, editors, {\em Advances in Neural Information Processing Systems}, volume~31. Curran Associates, Inc., 2018.

\bibitem{da_coarse_to_fine-Zheng-2020}
Y. Zheng, D. Huang, S. Liu, and Y. Wang.
\newblock Cross-domain object detection through coarse-to-fine feature adaptation.
\newblock In {\em CVPR}, 2020.

\bibitem{detic-zhou-2022}
X. Zhou, R. Girdhar, A. Joulin, and P.~Krähenbühl andd I.~Misra.
\newblock Detecting twenty-thousand classes using image-level supervision.
\newblock In {\em ECCV}, 2022.

\bibitem{selectivesearch}
Xinge Zhu, Jiangmiao Pang, Ceyuan Yang, Jianping Shi, and Dahua Lin.
\newblock Adapting object detectors via selective cross-domain alignment.
\newblock In {\em 2019 IEEE/CVF Conference on Computer Vision and Pattern Recognition (CVPR)}, pages 687--696, 2019.

\end{thebibliography}
